\documentclass[conference]{IEEEtran}
\IEEEoverridecommandlockouts
\usepackage{amsmath,amssymb,amsfonts}
\usepackage{algorithmic}
\usepackage{graphicx}
\usepackage{textcomp}
\usepackage{xcolor}
\usepackage{multirow}
\usepackage{makecell}
\usepackage{hyperref}
\usepackage[authoryear]{natbib}
\def\BibTeX{{\rm B\kern-.05em{\sc i\kern-.025em b}\kern-.08em
    T\kern-.1667em\lower.7ex\hbox{E}\kern-.125emX}}
\begin{document}

\title{SuperSAM: Crafting a SAM Supernetwork via Structured
Pruning and Unstructured Parameter Prioritization\\
\thanks{Github: \href{https://github.com/pnnl/SuperSAM}{Here is a link to the repository.}}
}

\author{%
  Waqwoya Abebe$^1$, Sadegh Jafari$^1$, Sixing Yu$^1$, Akash Dutta$^1$, \\
  Jan Strube$^2$, Nathan R. Tallent$^2$, Luanzheng Guo$^2$, \\
  Pablo Munoz$^3$, Ali Jannesari$^1$ \\[1em]
  $^1$Iowa State University, $^2$Pacific Northwest National Laboratory, \\
  $^3$Intel Labs \\[0.5em]
  \texttt{\{wmabebe, sadegh, yusx, adutta, jannesar\}@iastate.edu}, \\
  \texttt{\{jan.strube, nathan.tallent, lenny.guo\}@pnnl.gov}, \\
  \texttt{pablo.munoz@intel.com}
}

\maketitle

\begin{abstract}
  Neural Architecture Search (NAS) is a powerful approach of automating the design of efficient neural architectures. In contrast to traditional NAS methods, recently proposed one-shot NAS methods prove to be more efficient in performing NAS. One-shot NAS works by generating a singular weight-sharing supernetwork that acts as a search space (container) of subnetworks. Despite its achievements, designing the one-shot search space remains a major challenge. In this work we propose a search space design strategy for Vision Transformer (ViT)-based architectures. In particular, we convert the Segment Anything Model (SAM) into a weight-sharing supernetwork called SuperSAM. Our approach involves automating the search space design via layer-wise structured pruning and parameter prioritization. While the structured pruning applies probabilistic removal of certain transformer layers, parameter prioritization performs weight reordering and slicing of MLP-blocks in the remaining layers. We train supernetworks on several datasets using the sandwich rule. For deployment, we enhance subnetwork discovery by utilizing a program autotuner to identify efficient subnetworks within the search space. The resulting subnetworks are 30-70\% smaller in size compared to the original pre-trained SAM ViT-B, yet outperform the pretrained model. Our work introduces a new and effective method for ViT NAS search-space design.
\end{abstract}


\section{Introduction}

Vision Transformers (ViTs) \cite{dosovitskiy2020image} have transformed the landscape of computer vision by leveraging the self-attention mechanism originally developed for natural language processing. The Segment Anyting Model (SAM) \cite{SAM}, is a ViT-based foundation model for image segmentation. SAM was trained on the SA1B dataset \cite{SAM} which consists of 11M images and 1.1B mask annotations. SAM's model architecture is composed of three major components, a large ViT-based image encoder, a lightweight prompt encoder and a lightweight mask decoder. Given the size of image encoder and its computationally expensive attention mechanism, several works have been conducted in compressing the model for deployment in resource constrained environments \cite{zhang2023faster, xiong2023efficientsam, fu2024lite}.

Neural Architecture Search (NAS) techniques are advanced methods used to automatically discover efficient architectures for deep learning models. By automating the architectural design process, NAS methods aim to identify architectures that offer state-of-the-art performance while minimizing human intervention and computational cost. Recently, one-shot NAS methods have proven to be more efficient  compared to traditional NAS techniques \cite{white2023neural}. This is because one-shot NAS trains a single weight-sharing supernetwork that acts as a container of other subnetworks rather than train new architectures from scratch. As a result, the subnetworks directly inherit their weights from the same overparametrized supernetwork. This approach provides a means of training an exponential number of architectures for linear computation cost.

A major challenge in NAS involves designing of the the architecture search space \cite{munoz2024eftnas}. The search space is the set of all possible sub-architecture configurations (subnetworks) that could be derived by pruning the supernetwork (i.e. which is first initialized as the pre-trained SAM model). The supernetwork acts as an over-parameterized architecture that contains all possible sub architectures \cite{white2023neural}. Unsurprisingly, the search space is combinatorial making it quite infeasible to enumerate and optimize for the NAS task. In general, a NAS search space is first designed to bound the pool of potential architectures that can be derived. But even after bounding the search space, it could potentially contain millions of possible subnetwork configurations. Hence, an ideal search space not only reduces the number of candidate subnetworks but also contains promising subnetworks that require less resources to train. 

The NAS training process iteratively samples subnetworks from the search space and optimizes them on every batch of training data. In particular, our approach follows the `sandwich rule' \cite{yu2019universally}, where for a single batch of training data, the maximal subnetwork (supernetwork), the minimal subnetwork and a randomly selected subnetwork are sequentially sampled and optimized. After optimization, gradients of the sampled subnetworks are pushed back into the supernetwork (main model) for updating its parameters. 

Several search space design strategies have been proposed in the past. For instance, \cite{cai2019once} proposes Once-for-All, where 4-dimensional elasticity is applied to train a weight-sharing supernetwork for CNN architectures. They also propose progressive shrinking, a training strategy where the size of sampled subnetworks shrinks as the NAS training progresses. Recently, EFTNAS \cite{munoz2024eftnas} proposed performance aware search-space design coupled with first-order weight-reordering for transformer based models. NASViT\cite{gong2022nasvit} designs efficient small and medium-sized models. Its search space is inspired by LeViT\cite{graham2021levit}, which utilizes a hybrid architecture combining convolutions and transformers. For each CNN block, the search focuses on optimizing channel widths, block depths, expansion ratios, and kernel sizes. For each transformer block, it explores the optimal number of windows, hidden feature dimensions, depths, and MLP expansion ratios. The BigNAS\cite{bignas} search space includes multiple dimensions such as kernel sizes, channel numbers, input resolutions, and network depths. These dimensions are simultaneously searched to identify optimal child models.

In this work, we introduce SuperSAM, by transforming SAM into an `elastic' supernetwork, enabling the derivation of subnetworks with varying architectures tailored to a wide range of resource constraints. The subnetworks derived from SuperSAM exhibit comparable performance to the pre-trained SAM with just a fraction of its parameters. To do so, we propose combining probabilistic layer-wise pruning along side row/column-wise Wanda \cite{sun2023simple} parameter prioritization to design the search space and guide the subnetwork selection method. This approach first ranks the transformer layers and maintains the most important ones while the rest are assigned pruning probabilities. Moreover, the Wanda row/column-wise  prioritization performs row/column reordering and slicing of MLP blocks within remaining transformer layers to retain the most salient parameters. This proposed dual elasticity acts as a hierarchical mechanism where the structured pruning allows the discovery of a few high-performing subnetworks whereas the reordering and slicing operations expand the intermediate search space to discover a wide range of robust architectures.

In summary:

\begin{itemize}
    \item We propose a novel algorithm for search-space design for transformer-based NAS.
    \item We utilize this scheme to convert the SAM model into a weight-sharing supernetwork.
    \item We train supernetworks on multiple datasets demonstrating a higher training efficiency.
    \item We apply opentuner for subnetwork discovery and extract efficient subnetworks with comparable performance to the supernetwork.
\end{itemize}

The remainder of the paper is organized as follows: section \ref{sec:related_works} discusses related work, section \ref{sec:method}
 presents the methodology, section \ref{sec:evaluation} presents the evaluation. And we finally conclude the paper in section \ref{sec:conclusion}.

\section{Related works}
\label{sec:related_works}

\begin{figure*}
    \centering
    \includegraphics[width=0.8\textwidth, height=0.25\textheight]{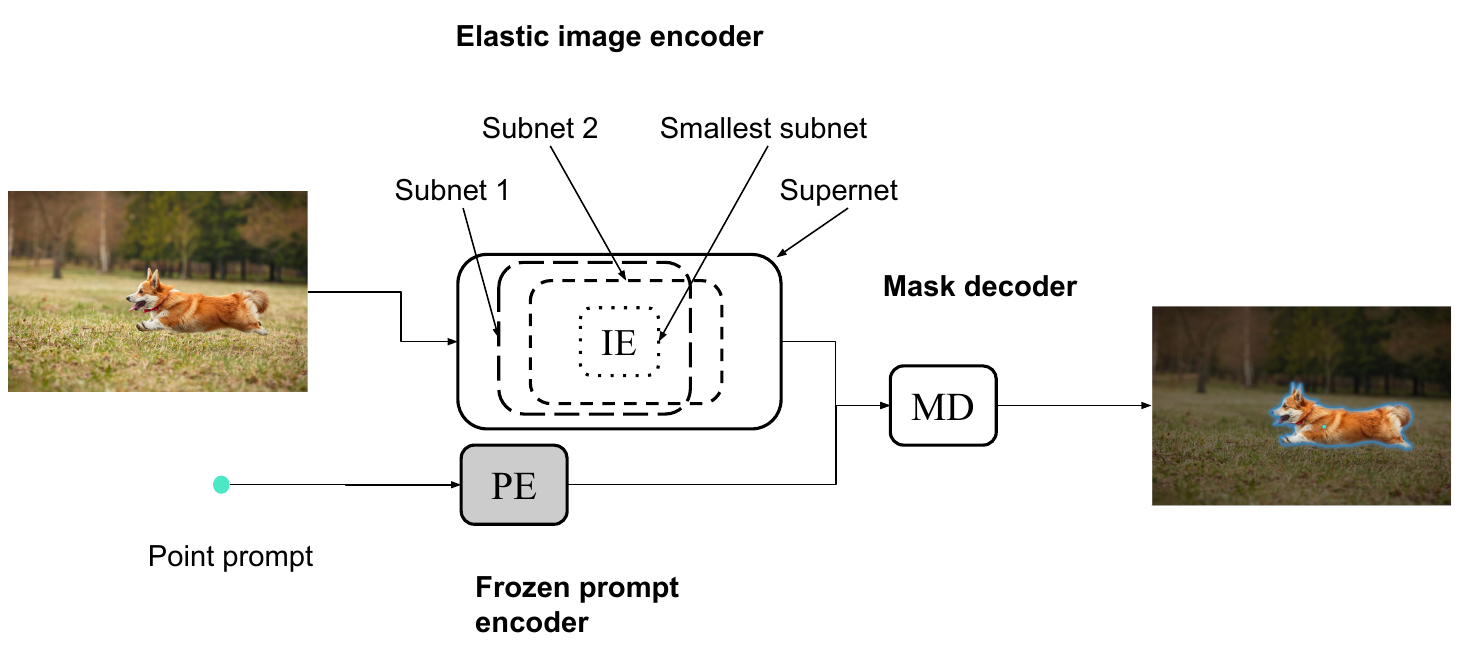} 
    \caption{After freezing the prompt encoder and applying 2D elasticity on the image encoder, the image encoder and mask decoder are jointly optimized using the sandwich rule. As shown above, while medium sized subnets may or may not overlap with each other, the smallest subnetwork lies in the intersectional region of all other subnetworks. }
    \label{fig:NAS_training}
\end{figure*}


\textbf{Neural Architecture Search:}

 The search space, consisting of a set of neural network architectures, is explored by NAS methods, which apply search and evaluation strategies to identify high-performing architectures that are often smaller and more efficient than human-crafted ones. \cite{elsken2019neural}
Traditional NAS methods require training each architecture from scratch, which is costly. In contrast, one-shot approaches\cite{white2023neural}, introduced in 2022, train all architectures in the search space simultaneously by using a single "supernetwork," an over-parameterized model containing all possible subnetworks. In one-shot weight-sharing approaches, it has been demonstrated through careful experimental analysis that it is possible to efficiently identify promising architectures from a complex search space \cite{bender2018understanding}. After training, search algorithms, such as reinforcement learning, evolutionary algorithms, or gradient-based methods, can be used to explore the vast space for possible architectures under certain constraints.

Various techniques exist for training the generated super-network. For instance, Once-for-All (OFA)\cite{cai2019once} introduces progressive shrinking, which enforces a training order that starts with large subnetworks and gradually moves to smaller ones. NASVIT\cite{gong2022nasvit} offers a set of methods, including a gradient projection algorithm, switchable layer scaling design, and a streamlined approach to data augmentation and regularization, all of which significantly enhance the convergence and performance of subnetworks. BigNAS \cite{bignas} challenges the conventional view that post-processing of weights is required for good prediction accuracy. Without additional retraining or post-processing, it trains a single set of shared weights. A sandwich sampling rule with inplace knowledge distillation (KD) is used to simultaneously optimize the supernet and sub-networks for each mini-batch, stabilizing training and improving convergence.
EFTNAS \cite{munoz2024eftnas} proposes using first-order weight reordering to improve the search space design. In particular, a column-wise reordering of attention and MLP layers is used to improve results of elasticity operation. 

\textbf{Network Pruning:}

Pruning is a popular technique for model compression. Generally, pruning can be categorized as structured or unstructured depending on the size of network components removed by the pruning operation. Structured pruning involves the removal of large network components\cite{cheng2024survey}. For example, in ShortGPT\cite{men2024shortgpt}, redundant layers are removed based on a metric that measures layer importance. LLM-Pruner\cite{ma2023llm} adopts a structured pruning approach by selectively removing non-critical coupled structures using gradient information. Similarly, BlockPruner\cite{zhong2024blockpruner} splits each transformer layer into Multi-Head Attention (MHA) and MLP blocks, evaluates their importance using perplexity measures, and employs a heuristic search for iterative pruning to optimize model efficiency. Additionally, Isomorphic Pruning\cite{fang2024isomorphic} demonstrates its effectiveness across various network architectures, such as Vision Transformers and CNNs, and consistently delivers competitive performance across models of different sizes, further showcasing the versatility of structured pruning techniques.

Unstructured pruning, also known as weight-wise pruning, targets individual weights by eliminating redundant connections in neural networks, setting the corresponding weights to zero. For example, magnitude pruning\cite{han2015learning} reduces storage and computation requirements by learning and retaining only the most important connections, achieving this reduction without sacrificing accuracy. Wanda \cite{Wanda} removes weights with the smallest magnitudes, adjusted by the corresponding input activations, on a per-output basis. Similarly, SparseGPT\cite{sparsegpt} approaches pruning as a layer-wise sparse regression problem, solving it approximately through a sequence of efficient Hessian updates and weight reconstructions, further optimizing neural network efficiency.\\

\textbf{SAM:}

Despite the impressive performance of the Segment Anything Model (SAM), its large ViT-based image encoder imposes a substantial inference cost, making it challenging to deploy in resource-constrained environments. Several methods have been proposed to address this issue. e.g., the Fast Segment Anything (FastSAM)\cite{zhao2023fast} introduces FastSAM, which employs YOLOv8-seg, an object detector adapted for instance segmentation, to significantly reduce computational overhead. MobileSAM\cite{zhang2023faster} was developed by applying decoupled knowledge distillation from ViT-H image encoder to a new tiny-ViT image encoder. EfficientSAMs\cite{xiong2023efficientsam} utilizes SAMI(masked image pretraining)  to enhance visual representation learning by reconstructing features from SAM’s image encoder. By combining SAMI-pretrained lightweight encoders with a mask decoder, EfficientSAMs achieve both efficiency and effectiveness. The SAM-Lightening\cite{songa2024SAM-Lightening} introduces Dilated Flash Attention, a re-engineered attention mechanism that increases inference speed by approximately 30 times compared to the original SAM.

\section{Methodology}
\label{sec:method}



\begin{figure*}
    \centering
    \includegraphics[width=0.8\textwidth]{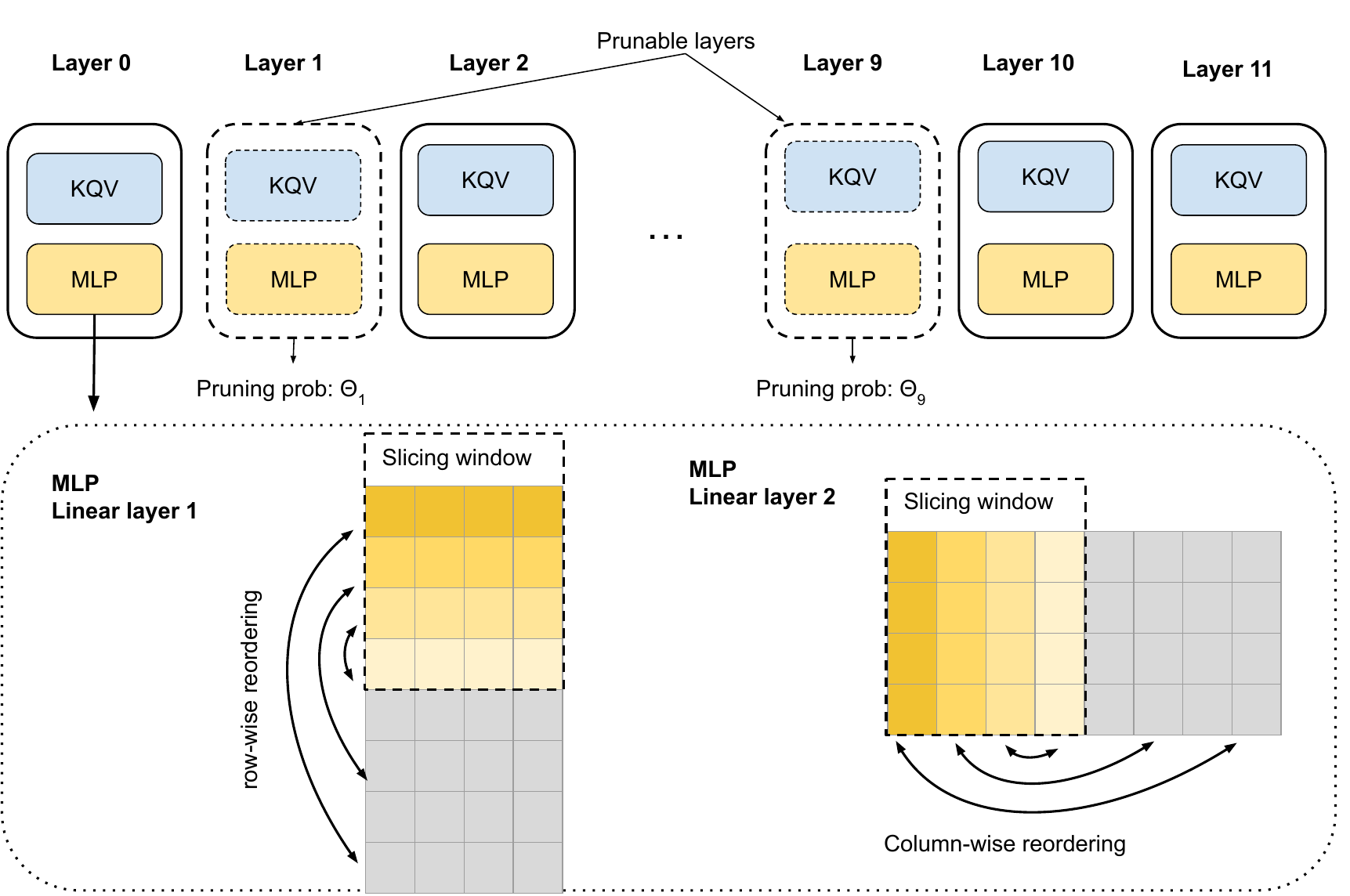} 
    \caption{The NAS search space design involves two main operations. 1. Identify prunable layers and assign a pruning probability. 2. Apply weight reordering in MLP blocks of surviving layers before a slicing window operation. These operations significantly reduce the size of the subnetworks in the search space as well as improve the quality of the subnetwork candidates. }
    \label{fig:elasticity}
\end{figure*}



\textbf{Search Space design:}

Following the intuition of previous compression attempts, we apply our proposed NAS technique on the image encoder. Unlike previous works that use direct distillation on a single architecture (eg. \cite{zhang2023faster}), we first apply elasticity to the image encoder and conduct NAS training. In NAS literature, elasticity describes the potential variability a certain architectural component can have across subnetworks \cite{munoz2021enabling}. For instance, two subnetworks ($\alpha^i$ and $\alpha^j$) derived from the same supernetwork could have varying number of channels for the same layer $k$, i.e. $\alpha_k^i \neq \alpha_k^j$. This makes layer $k$ elastic.

We apply 2-dimensional elasticity to the SAM ViT image encoder. The first elastic dimension is applied to the transformer layers as a form of structured pruning (similar to \cite{men2024shortgpt}). In this case, we utilize few-shot evaluation to determine the importance of the transformer layers with respect to a performance metric. This allows us to identify a set of prunable layers \(P \in \{0, 1, \ldots, 11\}\) that will be assigned a pruning probability $\theta_i$ during subnetwork selection. In particular, during NAS training, the pruning probability determines the selection of the corresponding transformer layer in the construction of the currently sampled subnetwork.

The second elasticity dimension applies weight reordering and slicing (windowing) of MLP-blocks in remaining (unpruned) layers. As shown in Fig. \ref{fig:elasticity}, MLP-blocks in remnant layers contain two linear layers. In the ViT-B version, these linear layers have a dimension of (3072 x 768) and (768 x 3072). Reordering and slicing is applied to the intermediate dimension (of size 3072). In particular, the rows of the first linear layer and the columns of the second linear layer are first ranked using the Wanda \cite{Wanda} metric. The row and column ranks are computed by aggregating Wanda scores of individual parameters in the rows and columns as shown in Eq. (\ref{eqn:reordering_score}) below:

\begin{equation}
    S_i = \sum_{i=0}^n {|W_{ij}| \cdot ||X_j||_2}
    \label{eqn:reordering_score}
\end{equation}

where $S_i$ is score for $i$-th column, $W_{ij}$ is the magnitude of the $ij$-th parameter and $||X_j||_2$ is the $l_2$ norm of the $j$-th input features aggregated across $N$ input batches with sequence length $L$.

\begin{figure}
    \centering
    \includegraphics[width=\columnwidth]{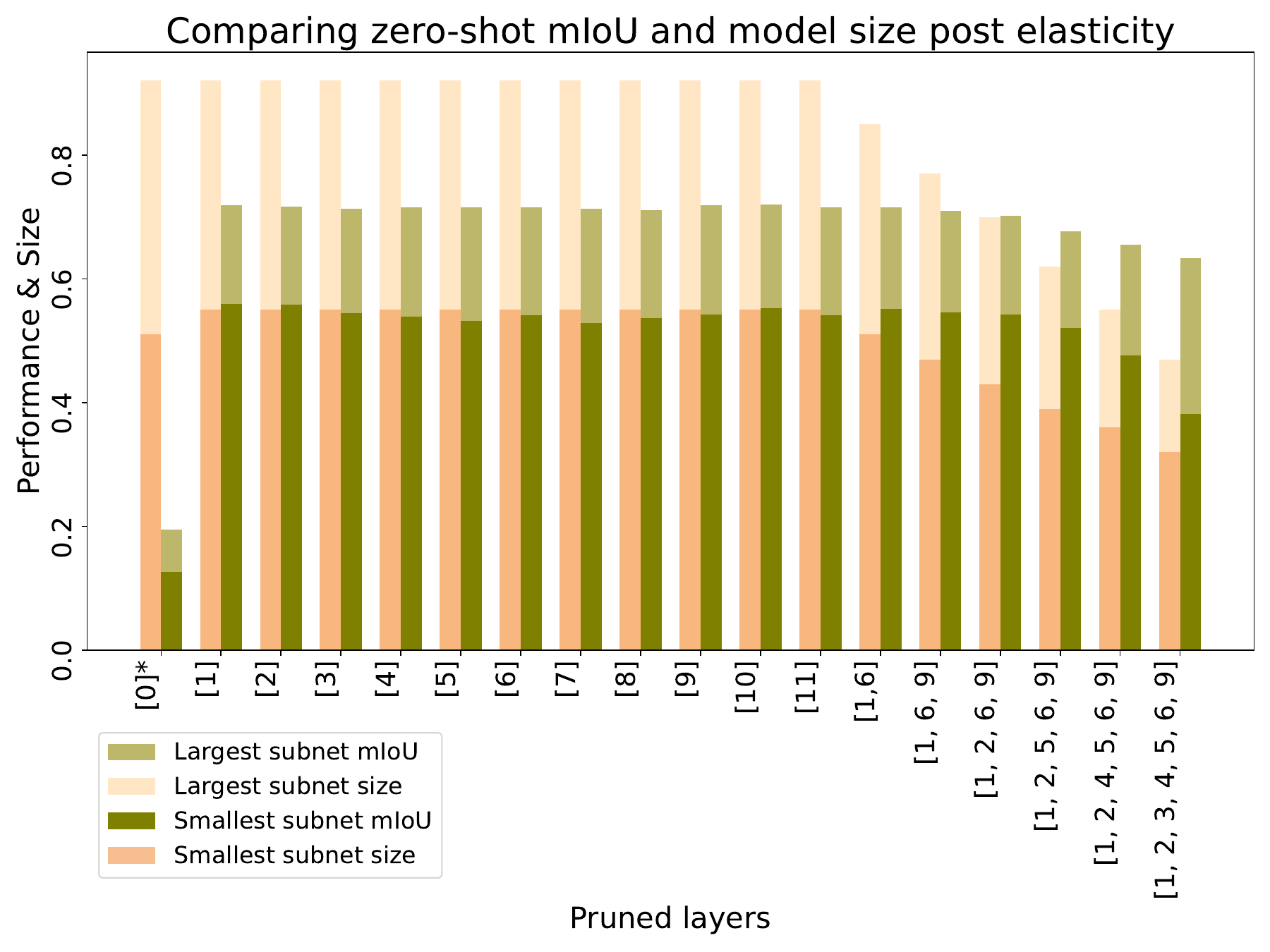}
    \caption{Comparing mIoU and model sizes by pruning one or more  layers from the SAM ViT-B image encoder. }
    \label{fig:layer_importance}
\end{figure}

Corresponding row and column ranks are averaged to compute a mean rank over the intermediate dimension (i.e. for both linear layers). For instance, the rank of the 3rd row in linear layer 1 is averaged with the rank of the 3rd column in linear layer 2 to compute the average rank of the third row-column pair. The rank averaging is computed across all corresponding row-column pairs to compute a mean rank over the intermediate dimension. The mean rank is then used to reorder corresponding rows and columns of linear layers 1 and 2 in descending order such that the reordered rows/columns align without distorting the forward/backward pass of the MLP-block. After reordering, a window \( w_i \; \text{for} \; i \in \{1, 2, \ldots, m\} \) is applied to the linear layers to slice the layers by retaining the first $w_i$  rows and columns. Note that slicing neither affects the dimension of input to linear layer 1 (i.e. 768), nor the output dimension of linear layer 2 (i.e. 768). Rather, it only shrinks the intermediate dimension (i.e. 3072). This approach effectively prioritizes and retains the most salient parameters in the linear layers significantly improving the quality of the search space. 

The proposed 2D elasticity is not arbitrary. Instead, we observe that the layer-wise pruning serves as starting points (anchors) in the search space whereas windowing operations expand on those anchors. Together, the layer-wise pruning and windowing act as a hierarchical mechanism for designing the search space. This is because, as previous work \cite{men2024shortgpt} has shown, and as we observed from preliminary experiments, transformer layers are not equally important. While some are critical, others can be redundant so much so that pruning (removing) layers may result in either a dramatic drop or marginal decrease in performance as shown in Fig. \ref{fig:layer_importance}. Layer pruning significantly reduces the model size facilitating the presence of a few high performing subnetworks. In contrast, when applying only windowing, the search space produces a substantial number of low-quality subnetworks. Nevertheless, we cannot solely depend  on layer-wise pruning, as this method results in only a limited set of high-quality subnets (anchors), typically numbering just a few dozen. As such, it poses a limitation in exploring a broader search space that includes smaller subnets dispersed between the anchors. This phenomenon is demonstrated in section \ref{sec:4.2}.

\textbf{NAS Training:} 

We implement the sandwich rule \cite{yu2019universally} to train the supernetworks. In particular, for each training batch, we sample the maximal subnet, the smallest subnet and a randomly selected subnet for sequential optimization. Here, we compute the cost using dice loss as shown in Eq. \ref{eq:dice} below.

\begin{equation}
\mathcal{L}_{\text{Dice}} = 1 - \frac{2 \times |GT \cap Pred|}{|GT| + |Pred|}
\label{eq:dice}
\end{equation}

where GT is ground truth mask and Pred is predicted mask.

In all three cases, we keep the prompt encoder and mask decoder architecturally intact across all subnetworks. i.e. Elasticity is only applied to the image encoder. This simplifies implementation and avoids slicing into smaller architectural components allowing all subnets to directly inherit the pre-trained architectures and weights. To train the SAM supernet, images are first resized to 1024 $\times$ 1024 and fed to the model along with prompts. SAM prompts can be one or more coordinate points and/or boxes on or around objects of interest \cite{kirillov2023segment}. We freeze the mask decoder during training and jointly optimize the image encoder and mask decoder as shown in Fig. \ref{fig:NAS_training}.

\textbf{Deployment:}

Once the SAM model is transformed into a supernetwork, its search space will have developed sufficiently to include subnetworks that perform on par with the supernetwork. As a result, a search algorithm can be used to identify subnets that meet specific resource constraints. For instance, we can query the search space to identify subnets with sizes under 60M parameters and mIoU performance $\geq$ 80\% on a given task.

Once SuperSAM is trained, therefore, we are now able to deploy opentuner \cite{opentuner}, a program auto-tuner to execute the constrained search. OpenTuner uses an
ensemble of disparate search techniques simultaneously; dynamically allocating a larger proportion of tests to 
 promising techniques. The ensemble techniques are themselves organized by a meta technique. For instance, an AUC Bandit meta technique can combine greedy mutation, differential evolution and hill climber instances to tackle the search problem. In the following section, we describe various experiments showcasing the efficacy of the proposed search space design strategy.

\section{Evaluation}
\label{sec:evaluation}





\begin{figure}
    \centering
    \includegraphics[width=\columnwidth]{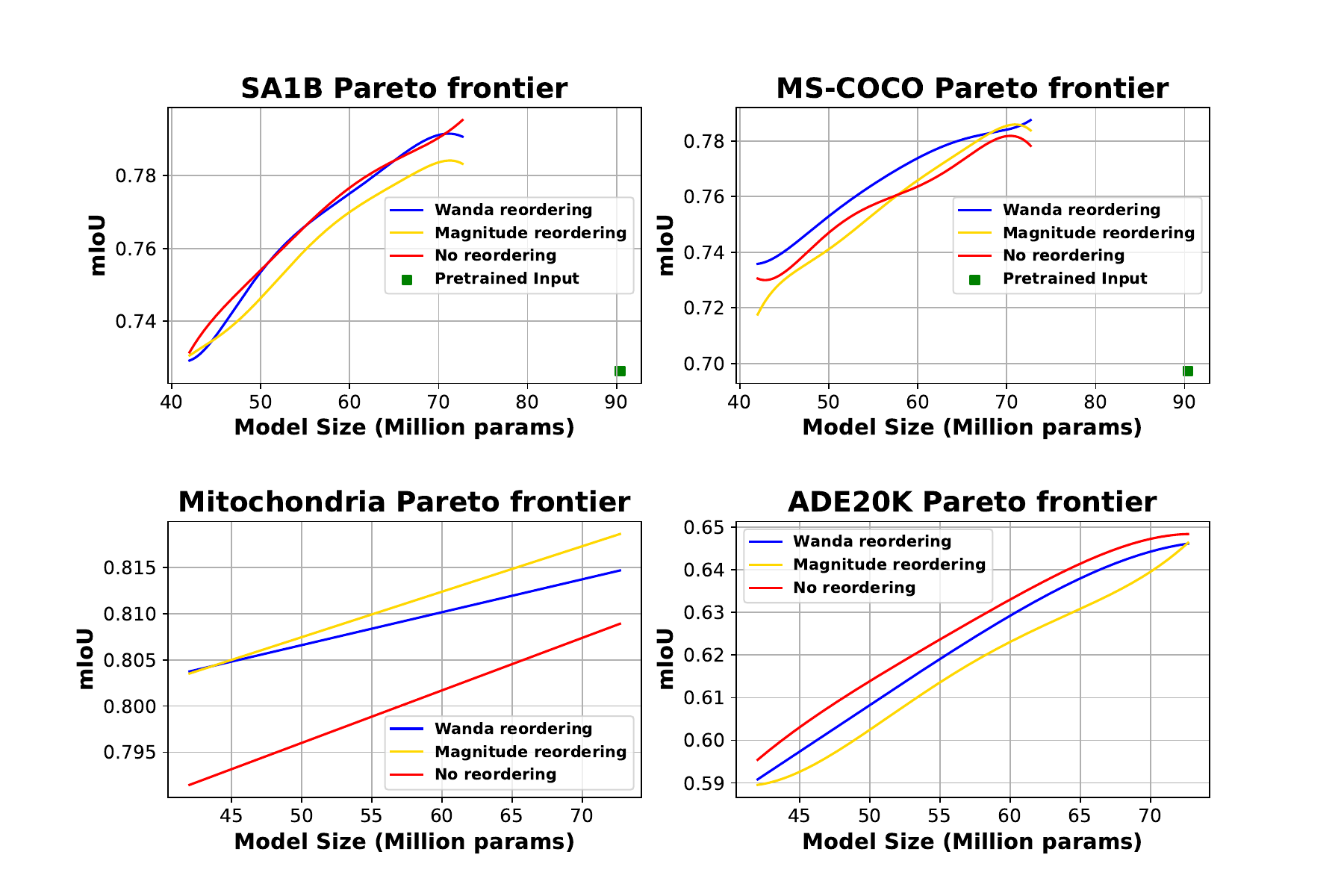}
    \caption{Comparing pareto frontier of different reordering strategies on different tasks. }
    \label{fig:pareto_comparison}
\end{figure}

\begin{figure}
    \centering
    \includegraphics[width=\columnwidth]{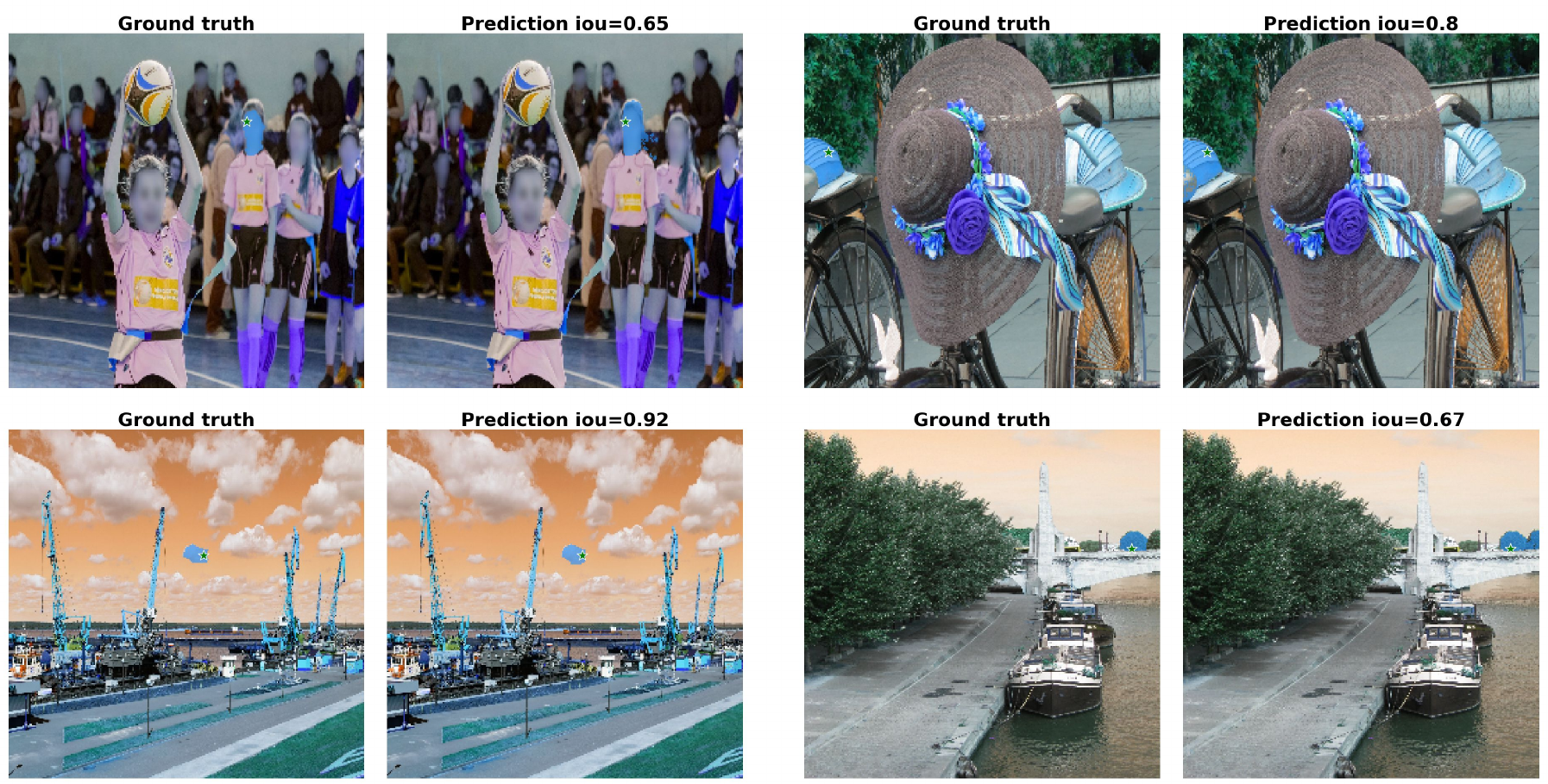}
    \caption{Sample data points selected from the SA1B validation set segmented using the smallest subnetwork (33.7M parameters) trained via proposed method. }
    \label{fig:sa1b_sample}
\end{figure}

\begin{figure}
    \centering
    \includegraphics[width=\columnwidth]{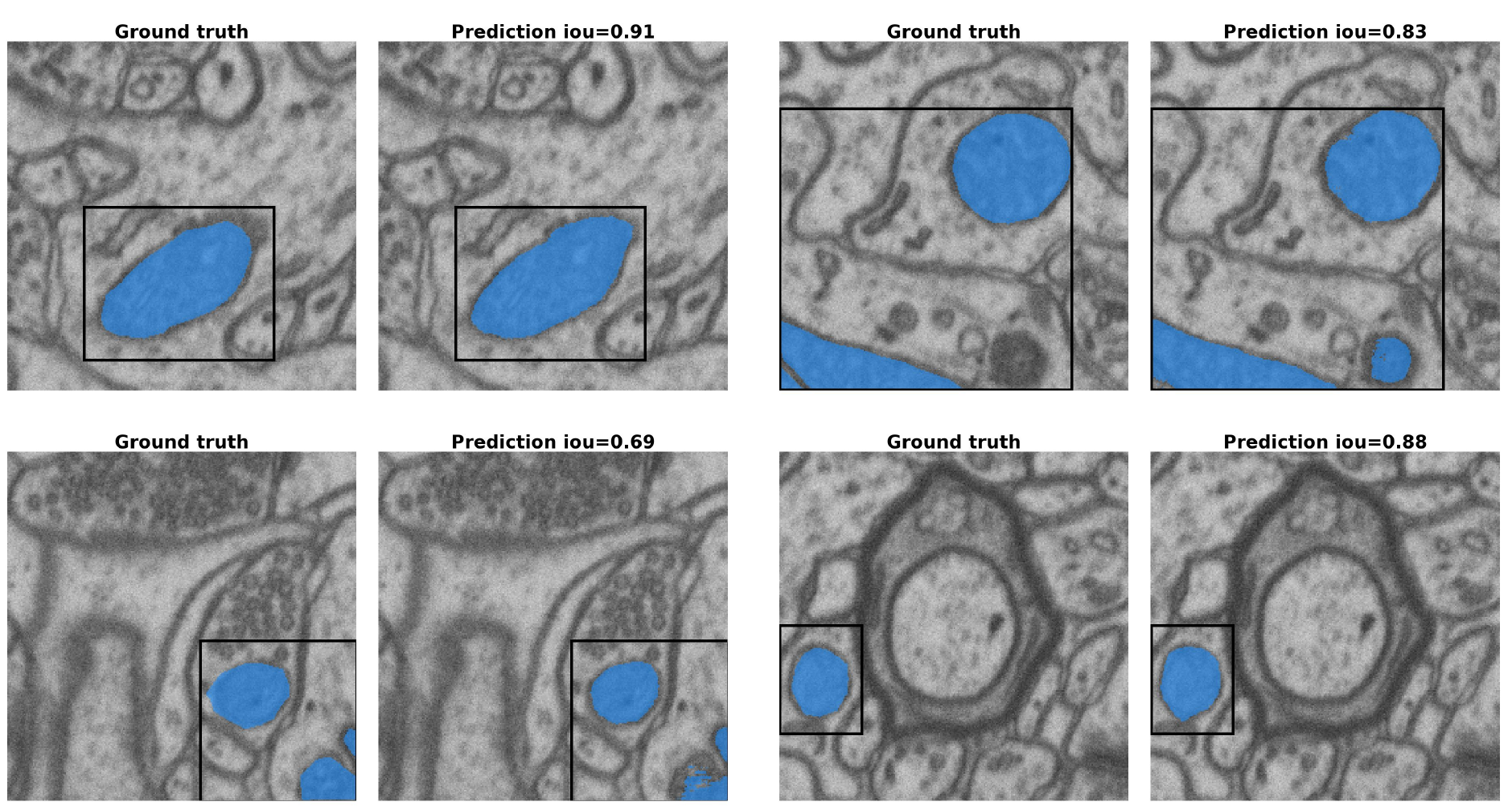}
    \caption{Sample data points selected from the Mitochondria validation set segmented using the smallest subnetwork (33.7M parameters) trained via proposed method. }
    \label{fig:mito_sample}
\end{figure}

We evaluate the proposed approach by training supernetworks using the SAM ViT-B model on different downstream tasks. These tasks include the SA1B \cite{kirillov2023segment} dataset that was used to train the SAM model, as well as MS-COCO \cite{lin2014microsoft}, ADE20K \cite{zhou2017scene} and Mitochondria \cite{lucchi2013learning} datasets. SA1B was used to train the SAM model, it contains around 11 million images and over 1.1 billion masks. In our experiments, we utilize 0.01\% of SA1B to train our supernetwork. i.e. 10,000 images for training and 100 for validation. We set a cutoff limit of 64 objects per image, by slicing the number of objects in the image if the objects exceed the cutoff limit and randomly repeating objects in case there are fewer objects.
 On the other hand, we utilize the entire MS-COCO dataset about 92K images for training and 5K for validation. Similar to the SA1B case, here, we set an object limit of 8 per image. Mitochondria is a small domain specific dataset containing gray-scale images of cells. We `patchify' the images into 256x256 patches to generate around 800 images for training and 800 for validation. Finally we use the entire train set, 20K images for training and a subset of 100 images for validation when using the ADE20K dataset.  We present example outputs of the smallest subnet in Figs. \ref{fig:sa1b_sample} and \ref{fig:mito_sample}.

 We train the supernets on the instance segmentation task  where either a single point prompt per object or bounding box is applied to segment the objects of interest. In particular, while we utilize box prompts for Mitochondria, we use point prompts for the other datasets. To account for prompt noise, we generate the prompts by selecting a random positive point inside the ground truth. Similarly, in the case of box prompts, we generate the box prompt around the object with a randomly chosen padding size of [0-20] pixels. We train our subnets using the DiceLoss \cite{sudre2017generalised} cost function and apply a learning rate of $1e-5$. We also apply a lambda learning rate decay until the learning rate shrinks to 1\% of its original value. 

In designing the search space, we use a fraction of the SA1B, 100 images, to compare the importance of transformer layers of the model. Similar to observations made by ShortGPT \cite{men2024shortgpt}, we notice that not all layers of the ViT-B image encoder are equally important. As shown in Fig \ref{fig:layer_importance}, layer 0 exhibits highest importance such that removing it causes a $\approx$60\% drop in mIoU. Moreover, after pruning layers [1,2,5,6,9] we notice that performance only drops by about $\approx$8\% whereas the model size has shrunk to 62\% of the original pre-trained model. With this insight, we design our search space by applying structured elasticity to layers [1,2,5,6,9]. Next, we apply column/row-wise elasticity windows of size [768, 1020, 1536, 2304], on remaining MLP blocks. Because of layer 0$^{th}$ importance, we keep it intact by shielding it from both layer-wise and row/column-wise elasticity operations.

\subsection{Reordering techniques}

\begin{table*}[ht]
\centering
\caption{Comparing computation cost for the smallest subnet to reach a target mIoU.}
\resizebox{\textwidth}{!}{%

\begin{tabular}{|cccccccc|}
\hline
\multirow{2}{*}{\textbf{Dataset (Task)}} & \multirow{2}{*}{\textbf{Target mIoU}} & \multirow{2}{*}{\textbf{\begin{tabular}[c]{@{}c@{}}Reordering technique\end{tabular}}} & \multirow{2}{*}{\textbf{Iterations}} & \multicolumn{3}{c}{\textbf{Training Cost}} \\
                                 &                                 &                                                                                     &                                   & \textbf{}            &  \textbf{Energy (KWhr)} & \textbf{Time (hrs.)} & \textbf{Carbon footprint (kg CO2e)}  \\ \hline

\multirow{3}{*}{\makecell{SA1B }}         &   \multirow{3}{*}{70\%}                      & No reordering                                                                             & 840    &                                                                & 5.35    &  7.68    & 2.34       \\
                                 &                        & Magnitude reordering                                                                                     & 1056                  &                                                                             & 6.73         & 9.66  & 2.94                 \\ 
                                 
                                 &                        & Wanda reordering                                                                                     &  672  &                                                                                           & \textbf{4.28}         & \textbf{6.14}  & \textbf{1.87}                 \\\hline
\multirow{3}{*}{\makecell{MS-COCO}}         &   \multirow{3}{*}{70\%}                      & No reordering                                                                             & 4312     &                                                                 & 10.12   & 8.12  & 4.42         \\
                                 &                        & Magnitude reordering                                                                                     &      5432       &                                                                                   & 12.75        & 12.23    & 5.57               \\ 
                                 &                        & Wanda reordering                                                                                     & 3192   &                                                                                        & \textbf{7.49}       & \textbf{6.01}    & \textbf{3.27}                 \\\hline
\multirow{3}{*}{\makecell{Mitochondria}}         &   \multirow{3}{*}{84\%}                      & No reordering                                                                             & 1000     &                                                                 & 0.95    & 10.10   & 0.42        \\
                                 &                        & Magnitude reordering                                                                                     &      1050       &                                                                                  & 0.99         & 10.60   & 0.43                \\ 
                                 &                        & Wanda reordering                                                                                     & 650   &                                                                                         & \textbf{0.61}         & \textbf{6.56}    & \textbf{0.27}                \\\hline                               
\multirow{3}{*}{\makecell{ADE20K}}         & \multirow{3}{*} {55\%}                      & No reordering                                                                             &  808  &                                                                 & 0.32    & 1.07     & 0.14         \\
                                 &                       &  Magnitude reordering                                                                                    &   1212
                                 &                                                                                   & 0.47         & 1.61      & 0.21               \\
                                 &                        & Wanda reordering                                                                                     &  606  &                                                                                           & \textbf{0.23}        & \textbf{0.81}     & \textbf{0.10}             \\\hline

\multicolumn{7}{l}{Carbon footprint computed using local emission factor 0.437 MT CO2e/MWh}

\end{tabular}%
}

\label{tab:com_cost}

\end{table*}

We start by training supernets on various tasks including, SA1B, MS-COCO, Mitochondria and ADE20K. In these experiments, we maintain identical layerwise elasticity settings and vary only the reordering techniques used to train the supernets. After training, we sample 100 subnetworks encompassing a wide range of sizes from the trained supernets. In particular, sampled architectures range in the size of 40M to 73M parameters in size compared to the supernet that has around 90M parameters.  In Fig. \ref{fig:pareto_comparison}, we compare the mIoU performance of the sampled architectures . For the SA1B and MS-COCO tasks, our proposed Wanda reordering strategy provides marginal performance gains over the ``no reordering" and magnitude reordering. Interestingly, the ``no reordering" performs slightly better on the ADE20K dataset, while it is outperformed on the domain specific mitochondria dataset. 

A significant challenge in NAS is the large amount of computational resources required to train the subnetworks. While one-shot NAS approaches help reduce the incurred costs, it remains expensive to train a supernet with a large search space. Therefore, it is imperative not just to train a supernet but also to consider the resources required to train it. In this regard, we compare the resources required to train the supernets to a given milestone, i.e. a target accuracy for the smallest subnet in the search space. Table \ref{tab:com_cost} shows that our proposed approach offers the fastest convergence on all tasks requiring fewer iterations to reach the target accuracy. Consequently, it requires less time and consumes less wattage to train.

\subsection{Effect of structured pruning}
\label{sec:4.2}

\begin{figure}
    \centering
    \includegraphics[width=\columnwidth]{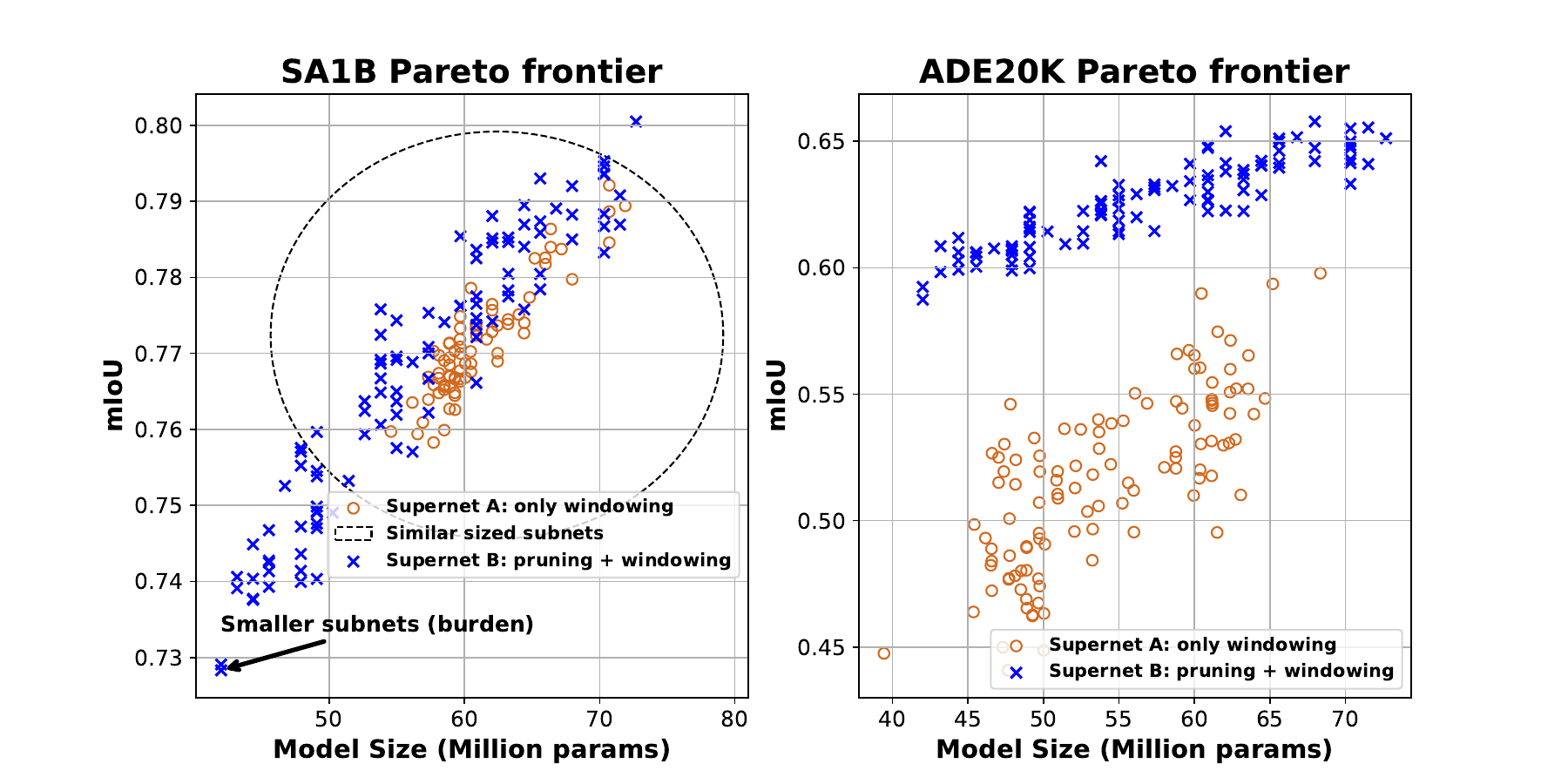}
    \caption{Comparing pareto frontier of different search space design strategies. Supernet A was generated via 1D elasticity (just windowing), Supernet B uses the proposed 2D elasticity. }
    \label{fig:pareto_comparison_2}
\end{figure}

We conducted an experiment to demonstrate the effect of structured pruning in designing the search space. In particular, we generated supernets on the SA1B and ADE20K tasks using two approaches. In the SA1B task, we designed the search space for the first supernet (A) using just windowing in all MLP-Blocks (except layer-0). For the second supernet (B), we applied the proposed method, i.e. structured pruning in conjunction with the same windowing strategy as in supernet A. As shown in Fig. \ref{fig:pareto_comparison_2}, Supernet B generates more robust subnetworks than Supernet A, despite comprising significantly smaller subnetworks that add to the training burden. Hence, despite the smallest subnet in B ($\approx$ 33.7M parameters) dragging its convergence, subenets in B exhibit superior performance than comparable subnets in A whose smallest subnet is $\approx$ 51.4M parameters in size. 

In the ADE20K task, we set out to implement a more fair comparison by involving smaller subnets in supernet A. We achieved this by adding an additional window $w_i$ for the MLP-Block [\textbf{256},768,1020,1536,2304], as well as adding a new window for the KQV-Blocks [\textbf{512},768]. This effectively reduced the smallest subnet in A to $\approx$ 33.8M parameters. In this scenario, we notice that supernet B has a significant performance edge over A as shown in Fig. \ref{fig:pareto_comparison_2}. This supports our intuition that structured pruning can enhance the design of the search space.

\section{Conclusion}
\label{sec:conclusion}

In this paper, we propose a new approach of search space design for training one-shot NAS in transformer-based vision models. We propose 2-dimensional elasticity where the first dimension applies probabilistic layer-wise structured pruning while the second dimension applies a Wanda-based row/column-wise reordering and slicing of MLP-Blocks of remaining layers. We show this hierarchical (2D) approach exploits the strengths of structured pruning to discover high performing subnetworks and utilizes windowing (slicing) to expand the search space to further discover robust subnetworks that lie in the intermediate space. We demonstrate the proposed technique on the Segment Anything Model (SAM), a newly developed foundation model for image segmentation. After training supernets on several tasks, we are able to discover and extract smaller networks with robust performance. In the SA1B case for instance, our subnetworks outperform the pretrained input model, SAM-ViT-B, despite the supernet being trained on just 0.1\% of the dataset. Moreover, our proposed Wanda-based reordering and windowing technique shows superior convergence compared to `no reordering', and `magnitude reordering' cases. This is quite important as one of the challenges of NAS is the computation requirements required to train subnetworks in the search space.

\section*{Acknowledgments}

This research is supported by the U.S.\@ Department of Energy (DOE) through
  the Office of Advanced Scientific Computing Research's ``Orchestration for Distributed \& Data-Intensive Scientific Exploration'' and
  the ``Cloud, HPC, and Edge for Science and Security'' LDRD at Pacific Northwest National Laboratory.
PNNL 
is operated by Battelle for the DOE under Contract DE-AC05-76RL01830.
This research is also supported by the National Science Foundation (NSF) under Award Number 2211982.
We thank Intel Labs and the Research IT team of Iowa State University for their support.

\bibliographystyle{plainnat}
\bibliography{references}

\end{document}